%% file: paper.tex
\documentclass[10pt,twocolumn,letterpaper]{article}

\usepackage{iccv}
\usepackage{times}
\usepackage{epsfig}
\usepackage{graphicx}
\usepackage{amsmath}
\usepackage{amssymb}
\usepackage[utf8]{inputenc}
\usepackage[percent]{overpic}
\usepackage{setspace}
\usepackage{balance}
\usepackage[pagebackref=true,breaklinks=true,letterpaper=true,colorlinks,bookmarks=false]{hyperref}

\iccvfinalcopy % *** Uncomment this line for the final submission

\ificcvfinal\fi
\begin{document}

%%%%%%%%% TITLE
\title{Improved Lossy Image Compression with Priming and Spatially Adaptive
  Bit Rates for Recurrent Networks}

\ificcvfinal
\author{
  Nick Johnston \\
  Google Inc. \\
  {\tt\small{nickj@google.com}} \\
  \and
  Damien Vincent \\
  {\tt\small{damienv@google.com}} \\
  \and
  David Minnen \\
  {\tt\small{dminnen@google.com}} \\
  \and
  Michele Covell \\
  {\tt\small{covell@google.com}} \\
  \and
  Saurabh Singh \\
  {\tt\small{saurabhsingh@google.com}} \\
  \and
  Troy Chinen \\
  {\tt\small{tchinen@google.com}} \\
  \and
  Sung Jin Hwang \\
  {\tt\small{sjhwang@google.com}} \\
  \and
  Joel Shor \\
  {\tt\small{joelshor@google.com}} \\
  \and
  George Toderici \\
  {\tt\small{gtoderici@google.com}} \\
}
\else
\author{First Author\\
Institution1\\
Institution1 address\\
{\tt\small firstauthor@i1.org}
\and
Second Author\\
Institution2\\
First line of institution2 address\\
{\tt\small secondauthor@i2.org}
}
\fi
\maketitle
%\thispagestyle{empty}

%%%%%%%%% ABSTRACT
\begin{abstract}
  \input{abstract.tex}
\end{abstract}

%%%%%%%%% BODY TEXT
\input{intro.tex}
\input{related_work.tex}

\input{methods.tex}

\input{training.tex}
\input{results.tex}
\input{conclusion.tex}
%-------------------------------------------------------------------------

%\pagebreak

\balance

{\small
\bibliographystyle{ieee}
\balance
\bibliography{egbib.bib}
}

\end{document}

%% file: abstract.tex
We propose a method for lossy image compression based on recurrent,
convolutional neural networks that outperforms BPG (4:2:0), WebP, JPEG2000,
and JPEG as measured by MS-SSIM. We introduce three improvements over
previous research that lead to this state-of-the-art result. First, we show
that training with a pixel-wise loss weighted by SSIM increases
reconstruction quality according to several metrics. Second, we modify the
recurrent architecture to improve spatial diffusion, which allows the
network to more effectively capture and propagate image information through
the network's hidden state. Finally, in addition to lossless entropy coding,
we use a spatially adaptive bit allocation algorithm to more efficiently use
the limited number of bits to encode visually complex image regions. We
evaluate our method on the Kodak and Tecnick image sets and compare against
standard codecs as well recently published methods based on deep neural
networks.

%% file: intro.tex
%\vspace{-0.5em}
\section{Introduction}
%\vspace{-0.5em}
\label{section: intro}

Previous research showed that deep neural networks can be effectively applied
to the problem of lossy image compression~\cite{TwitterComp, Toderici2016iclr,
  Toderici2017cvpr, ConceptualComp, Balle2017iclr}. Those methods extend the
basic autoencoder structure and generate a binary representation for an image
by quantizing either the bottleneck layer or the corresponding latent
variables. Several options have been explored for encoding images at different
bit rates including training multiple models~\cite{Balle2017iclr}, learning
quantization-scaling parameters~\cite{TwitterComp}, and transmitting a subset
of the encoded representation within a recurrent
structure~\cite{ConceptualComp, Toderici2017cvpr}.

Our method takes the recurrent approach and builds on the architecture
introduced by~\cite{Toderici2017cvpr}. The model uses a recurrent autoencoder
where each iteration encodes the residual between the previous reconstruction
and the original image (see~\autoref{fig: single-iter-gru-1021}). At each
step, the network extracts new information from the current residual and
combines it with context stored in the hidden state of the recurrent
layers. By saving the bits from the quantized bottleneck after each iteration,
the model generates a progressive encoding of the input image. Low bit rate
encodings then use the codes from early iterations, while reconstructions at
higher bit rates also use the codes from later iterations.

\begin{figure}
  \centering \setstretch{0.95}
   \begin{overpic}[width=\linewidth]{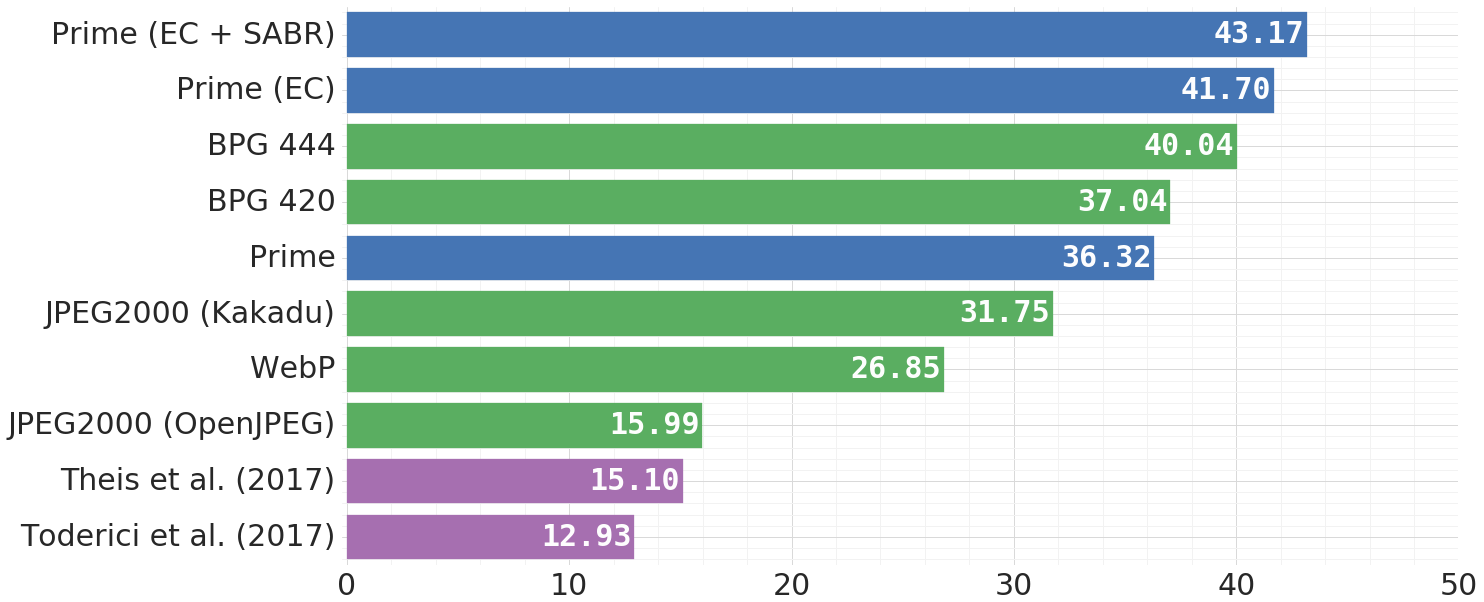}
    \put (85,5) {Kodak}
  \end{overpic}
  \caption{\small Our approach outperforms standard codecs and existing
    neural-network-based methods. This figure shows rate savings
    (Bj{\o}ntegaard Delta) relative to JPEG under MS-SSIM for the Kodak
    dataset. Standard codecs are shown in green, purple represents recent
    research using neural networks~\cite{TwitterComp, Toderici2017cvpr}, and
    our methods are shown in blue.}
  \label{fig: bjontegaard_kodak_rate}
\end{figure}

%% Image compression is fundamentally concerned with removing spatial correlation
%% to create a more compact representation of an image. In the decoder portion of
%% our model, spatial context is provided by convolutional kernels that extract
%% features from both the input (either the binary representation or features
%% maps derived from it) and, for recurrent layers, the hidden state. This
%% context allows the network to exploit correlations across space with the
%% extent of the spatial context growing as the number of recurrent steps
%% increases.

Our method provides a significant increase in compression performance over
previous models due to three improvements. First, by ``priming'' the network,
that is, running several iterations before generating the binary codes (in the
encoder) or a reconstructed image (in the decoder), we expand the spatial
context available to the decoder and allow the network to represent more
complex functions in early iterations. Second, we add support for spatially
adaptive bit rates (SABR), which dynamically adjusts the bit rate across each
image depending on the complexity of the local image content. Finally, we
train our model with a more sophisticated loss function that guides the
pixel-wise loss using structural similarity (SSIM)~\cite{ssim}. As summarized in
\autoref{fig: bjontegaard_kodak_rate}, adding these three techniques yields a
rate-distortion (RD) curve that exceeds state-of-the-art codecs (BPG 444 (YCbCr
4:4:4)~\cite{BPG}, BPG 420 (YCbCr 4:2:0), WebP~\cite{WebP},
JPEG2000~\cite{jpeg2000}, and JPEG~\cite{JPEG}) as well as other learned
models based on deep neural networks (\cite{TwitterComp}
and~\cite{Toderici2017cvpr}), as measured by MS-SSIM~\cite{wang2003multiscale}.

We review previous work in~\autoref{section: related} and describe our method
in detail in~\autoref{section: methods}. The description focuses on the
network architecture (\autoref{subsection: architecture}), how we combine that
with hidden-state priming and diffusion (\autoref{subsection: priming}), and
how we use spatially adaptive bit rates (\autoref{subsection: sabr}).
\autoref{section: methods} also covers our training loss (\autoref{subsection:
  ssim}), which provides better generalization results than unweighted $L_1$
or $L_2$ loss. \autoref{section: results} summarizes the results and compares
them to existing codecs and to other recent research in neural-network-based
compression~\cite{TwitterComp}.

%% file: related_work.tex
%\vspace{-0.5em}
\section{Related Work}
%\vspace{-0.5em}
\label{section: related}
%There are two main categories of compression algorithms: lossless and lossy.
%Entropy of the data distribution dictates the maximum possible lossless
%compression achievable limiting its application. Lossy compression allows a

%Nearly all image compression currently used is lossy, making it a
%tradeoff between the quality and compression rate.

%making it suitable for
%several image compression. Lossy image compression has been widely
%studied with

%% The JPEG~\cite{JPEG} standard was first released in 1992 and remains the most
%% widely used method for lossy compression of digital
%% photographs~\cite{bull2014book}. The Joint Photographic Experts Group released
%% JPEG2000 eight years later with support for additional features such as larger
%% resolutions, region of interest coding, data security, and higher coding
%% efficiency~\cite{jpeg2000}. Released in 2010, WebP is based on the intra-frame
%% coding method used by the VP8 video codec with the goal of further increasing
%% coding efficiency and improving animation and transparency support for images
%% on the web~\cite{WebP}. Finally, Better Portable Graphics (BPG) was released
%% in 2014 and is based on the intra-frame coding method used by the High
%% Efficiency Video Coding (HEVC) standard~\cite{BPG}. To our knowledge, BPG
%% currently has the highest coding efficiency for lossy image compression
%% amongst public codecs.

Lossy image compression is a long-standing problem in image processing and
many standard codecs have been developed. JPEG~\cite{JPEG} remains the most
widely used method for lossy compression of digital
photographs~\cite{bull2014book}. Several more sophisticated standards were
developed including JPEG2000~\cite{jpeg2000}, WebP~\cite{WebP}, and Better
Portable Graphics (BPG)~\cite{BPG}. To our knowledge, BPG currently has the
highest coding efficiency for lossy image compression amongst public codecs.

Recently, there has been a surge in research applying neural networks to the
problem of image compression~\cite{TwitterComp, Toderici2016iclr,
Toderici2017cvpr, ConceptualComp, Balle2017iclr, Santurkar2017gencomp,
Baig2017colorization}. While such methods were explored since at least the
late 1980s~\cite{munro1989image, Jiang1999}, few neural-network-based systems
improve upon JPEG or match the coding efficiency of JPEG2000 on typical
digital images found on the web.

% Early work in neural networks considered their use for image compression in the .
% Recent advances in deep learning in conjunction with powerful hardware are
% now enabling us to revisit these ideas.
%The recent interest in neural networks based compression approach can be
%attributed to two main reasons: 1) Deep neural networks have been very
%succesful in various computer vision applications including
%classification~\cite{he2016resnet,szegedy2016inception},
%object detection~\cite{szegedy2013deep,FRCNN,redmon2016you},
%segmentation~\cite{deeplab,liu2015parsenet,badrinarayanan2015segnet},
%pose estimation~\cite{wei2016convolutional,tompson2014joint,cao2016realtime}
%among numerous others, where they discover and exploit
%latent patterns in data, 2) They
% Neural-network-based approaches are of interest since they
% allow the encoder and decoder to be learnt
% from data,
%making it trivial to adapt them to specific domains and application needs.
% possibly discovering patterns and transforms that are best matched to
% an application domain or a quality metric.

% used for dimensionality reduction and
Autoencoders with a bottleneck have been used to learn compact representations
for many applications~\cite{Hinton2006, Krizhevsky2011, vincent2010} and form
the basis for most network-based compression models. Theis {\it et al.} used
an ensemble of encoders and target multiple bit rates by learning a scaling
parameter that changes the effective quantization granularity. Ball\'{e} {\it
et al.} use a similar architecture but use a form of local gain control called
generalized divisive normalization~\cite{Balle2016iclr} and replace the
non-differentiable quantization step with a continuous relaxation by adding
uniform noise~\cite{Balle2017iclr}.

%% Each of the different members of the ensemble of encoders is trained for a
%% different target bit-rate range (``low'', ``medium'', or ``high'') but the
%% actual decision on which encoder/decoder pair to use is made based on the
%% observed reconstruction qualities.

A different method for targeting multiple bit rates uses recurrent
autoencoders~\cite{ConceptualComp,Toderici2016iclr,Toderici2017cvpr}. In this
approach, the model generates a progressive encoding that grows with the
number of recurrent iterations. Different bit rates are achieved by
transmitting only a subset (prefix) of the progressive code. Gregor {\it et
al.} use a generative model so missing codes are replaced by sampling from the
learned distribution~\cite{ConceptualComp}. Our model uses a recurrent
architecture similar to Toderici {\it et al.} where missing codes are
ignored~\cite{Toderici2017cvpr}. The decoder thus runs fewer iterations for
low bit rate encodings and will generate a valid, but less accurate,
reconstruction compared to high bit rate encodings.

% through the network: the encoder refines the representation on each recursion,
% by adding more bits (taken from the bottleneck layer) on each pass. 

% TODO(dminnen) adaptive computation research is conceptually related but not
% directly related to our work. Priming / diffusion use a fixed (per model)
% amount of computation (whereas ACT is dynamic), and SABR doesn't learn to be
% dynamic.

%% Adaptive computation has been proposed for early termination and saving
%% computation~\cite{graves2016adaptive,figurnov2016spatially}. Our
%% method uses similar techniques to instead improve our final
%% rate-distortion performance.

%doesn't explicitly learn to minimize computation, instead we study the effect of
%varying number of priming and diffusion steps on the final performance.

%% file: methods.tex
\begin{figure}
  \begin{minipage}{\columnwidth}
  \centering \setstretch{0.95}
  \includegraphics[width=0.9\linewidth]{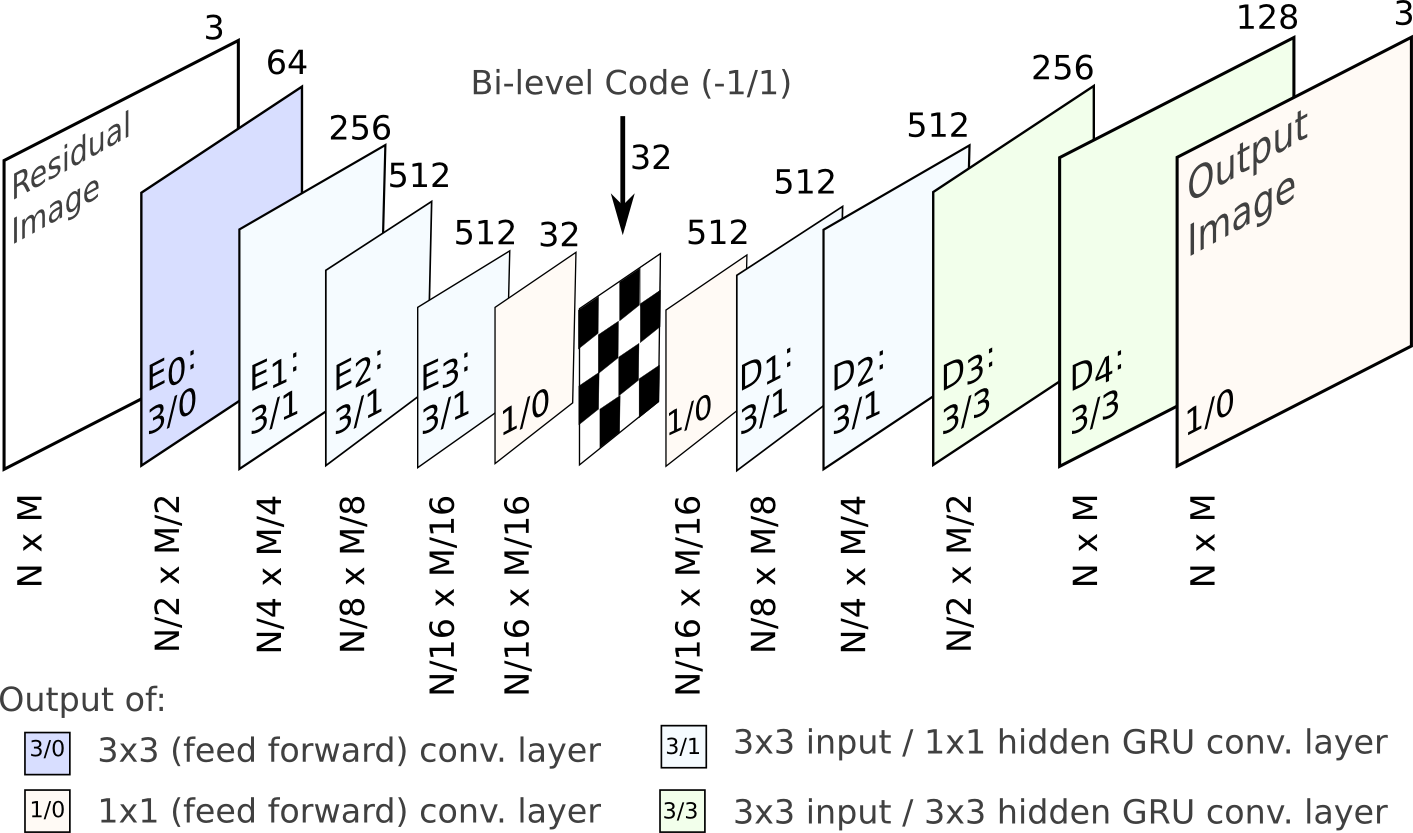}
  \caption{\small The layers in our compression network, showing the encoder
    ($E_i$), binarizer (checkerboard), and decoder ($D_j$). Each layer is
    labeled with its relative resolution (below) and depth (above). The inner
    label (``I / H'') represents the size of the convolutional kernels used for
    input (I) and for the hidden state (H).}
  \label{fig: single-iter-gru-1021}
  \end{minipage}
  \begin{minipage}{\columnwidth}
  \centering \setstretch{0.95}
  \vspace{2em}
  \includegraphics[width=\linewidth]{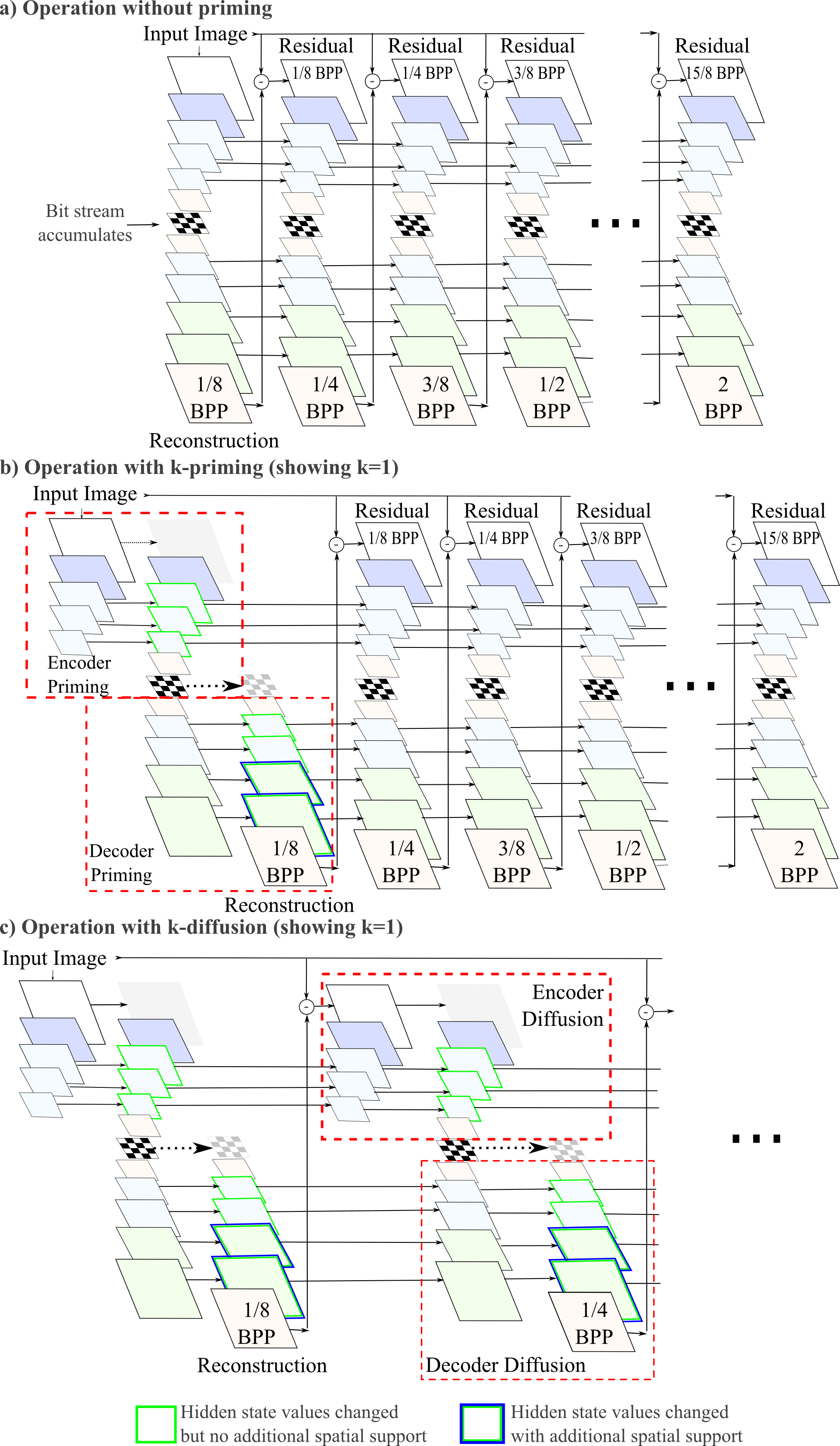}
  \caption{\small Network operation: (a) without priming, (b) with priming,
    and (c) with diffusion.}
  \label{fig: priming arch}
  \vspace{-2em}
  \end{minipage}
\end{figure}

%\vspace{-0.5em}
\section{Methods}
%\vspace{-0.5em}
\label{section: methods}
In this section, we first describe the network architecture used in our
research along with an analysis of its spatial support. We then describe each
of the three techniques that we leverage to achieve our results:
perceptually-weighted training loss, hidden-state priming, and spatially
adaptive bit rates.

\subsection{Network Architecture}
\label{subsection: architecture}

\autoref{fig: single-iter-gru-1021} shows the architecture used for our
encoder and decoder networks.  The depth of each layer is marked above the
back corner of each plane. The name and type of layer is depicted as ``$E_i :
I / H$'' for the encoder (and ``$D_j : I / H$'' for the decoder) inside the
bottom of each plane. The convolutional kernels for input have size $I \times
I$ and the convolutional kernels for the hidden state are $H \times
H$. Feed-forward, non-recurrent layers have $H=0$. The input to the encoder is
the residual image: the difference between the original image and previous
iteration's reconstruction. For the first iteration, this residual is simply
the original image.

The first and last layers on both the encoder and decoder networks use
feed-forward convolutional units ($H=0$) with $\tanh$ activations. The other
layers contain convolutional Gated Recurrent Units
(GRU)~\cite{chung2014empirical}.

To ensure
accurate bit-rate counts, the binarizer (shown as a checkerboard in
\autoref{fig: single-iter-gru-1021}) quantizes its input to be
$\pm 1$~\cite{Toderici2016iclr}.
This will give us our nominal (pre-entropy coding) bit
rate.  Given our choice of downsampling rates and binarizer depths,
each iteration adds $\frac{1}{8}$ bpp to the previous nominal bit rate.

The spatial context used by each reconstruction pixel,
as a function of either the ``bit stacks'' (that is, the outputs of the binarizer
at a single spatial position)
or the original image pixels,
can be computed by examining
the combined spatial supports of the encoder, the decoder, and all state
vectors.\footnote{Detailed derivations of these equations, as well as the ones for
  the priming and diffusion supports, are given in the Supplementary Material.}
The dependence of the output reconstruction on the bit
stacks varies by output-pixel position by one
bit stack (in each spatial dimension), so we will discuss only
the maximum spatial support:
\vspace{-0.5em}
{\setstretch{0.95}
\begin{align}
  \max({\cal S}_B(F_t)) & = 6 t + \left\lceil{5.5}\right\rceil
  \label{eq: S_B(F)} \\
{\cal S}_I(F_t) & = 16 {\cal S}_B(F_t) + 15
  \label{eq: S_I(F)}
\end{align}}
\noindent where ${\cal S}_B(F_t) \times {\cal S}_B(F_t)$ and
${\cal S}_I(F_t) \times {\cal S}_I(F_t)$ are the spatial support of
the reconstruction on the bit stacks and on the original image
pixels, respectively.

\subsection{Hidden-state Priming}
\label{subsection: priming}

\begin{figure}
  \begin{minipage}{\columnwidth}
  \centering \setstretch{0.95}
  \includegraphics[width=0.325\linewidth]{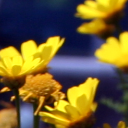}
  \hfill
  \includegraphics[width=0.325\linewidth]{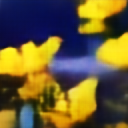}
  \hfill
  \includegraphics[width=0.325\linewidth]{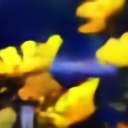}
  \caption{\small Left: Crop of the original Tecnick image 98.
    Center: Reconstruction using the DSSIM network at 0.125 bpp.
    Right: Reconstruction using the Prime network at 0.125 bpp.
    Notice the reduced artifacts from right versus center, especially in the
    sunflower in the lower left corner.}
  \label{fig: prime_crop}
  \end{minipage}
  \\ ~ \\ ~ \\
  \begin{minipage}{\columnwidth}
  \centering \setstretch{0.95}
  \includegraphics[width=0.24\linewidth]{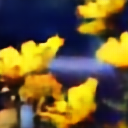}
  \hfill
  \includegraphics[width=0.24\linewidth]{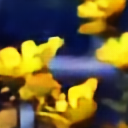}
  \hfill
  \includegraphics[width=0.24\linewidth]{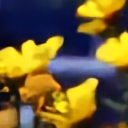}
  \hfill
  \includegraphics[width=0.24\linewidth]{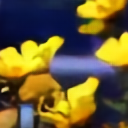}
  \caption{\small Cropped reconstructions of Tecnick image 98, taken at 0.25 bpp.
    From left to right, the results are from
    networks with no diffusion (0-diffusion) up to 3-diffusion.
    Notice the increased petal definition as more diffusion is used.}
  \label{fig: abp_crop}
  \end{minipage}
\end{figure}

On the first iteration of our compression networks, the hidden states
of each GRU layer are initialized to zero (\autoref{fig: priming arch}-a).
In our experiments, we have seen a strong
visual improvement in image quality across the first several iterations. Our
hypothesis is that not having a good hidden-state
initialization degrades our early bit-rate performance.
Since both encoder and decoder architectures stack several GRU layers
sequentially, it takes several iterations for the hidden-state improvement
from the first GRU layer to be observable at the binarizer (for the encoder)
or in the reconstruction (for the decoder).
% and increasing the
%spatial extent leads to increased reconstruction quality.
Our approach to tackling this problem is to generate a better initial
hidden-state for each layer with a technique called hidden-state
priming.

Hidden-state priming, or ``$k$-priming'', increases the recurrent-depth of
the first iteration of the encoder and decoder networks,
separately, by an additional $k$ steps (\autoref{fig: priming
arch}-b). To avoid using additional
bandwidth, we run these additional steps separately, without
adding the extra bits produced by the encoder to the actual bit stream.
%we can't allow additional information to
%flow across the binarizer, as this would require transmitting
%additional information at decode time.
For the encoder, this means
processing the original image multiple times, while discarding
the generated bits but keeping the changes to the hidden state within the encoder's
recurrent units. For the decoder, this means taking the first valid
set of bits transmitted and generating a decoded image multiple
times but only keeping
the final image reconstruction (and the changes to the decoder's hidden states). \autoref{fig:
prime_crop} depicts an example patch of an image from our evaluation
set and the reconstructions from our networks trained with and without
priming.  The reconstruction with priming is
both visually and quantitatively better than without priming, without
using any additional bits.

Priming can be performed between iterations as
well. We call this diffusion (\autoref{fig: priming
arch}-c). Diffusion has experimentally shown
better results (\autoref{fig: abp_crop}), but at the cost
of runtime and training time. When $k$ steps are added in between each
emitting iteration, we call this $k$-diffusion. As we increase $k$, we
both increase the maximum support of the system along with computation
and training time.

In addition to achieving a better hidden-state representation for our
networks, priming and diffusion also increase the spatial extent of
the hidden-states, in the decoder, where the last two layers of the
hidden kernels are $3 \times 3$, and in later iterations of the
encoder, when the increased decoder support propagates to increased
encoder support. This changes $\max({\cal S}_B(F_t))$ from
\autoref{eq: S_B(F)} to
\vspace{-0.5em}
$$\max({\cal S}_B(F_t)) = \left\lceil{1.5 k_d + 5.5}\right\rceil t + \left\lceil{1.5 k_p + 5.5}\right\rceil$$
with $k_p = k_d$ when $k_d > 0$.

\subsection{Spatially Adaptive Bit Rates}
\label{subsection: sabr}

By construction, our recurrent models generate image representations at
different bit rates according to the number of iterations, but those bit rates
are constant across each image. That means that the local (nominal)
bit rate is fixed
regardless of the complexity of the underlying image content, which is
inefficient in terms of quantitative and perceptual quality (e.g., consider
the number of bits needed to accurately encode a clear
%, blue
sky compared to a flower bed).
%human face)
%or highly textured forest scene).

In practice, the entropy coder introduces some spatial adaptivity based on the
complexity and predictability of the binary codes, but our training procedure
does not directly encourage the encoder to generate low-entropy
codes. Instead, the loss function only pushes the network to maximize
reconstruction quality over image patches. In order to maximize quality across
a full image for a target (average) bit rate, we introduce a spatially
adaptive bit rate (SABR) post-process to dynamically adjust the local bit
rate according to a target reconstruction quality.

The results presented in \autoref{section: results} use a very simple bit
allocation algorithm. %, though a more sophisticated method can be easily
%substituted.
Given a target quality, each image tile is assigned as many bits
as necessary to meet or exceed the target quality up to the maximum supported
by the model.

Since our decoder uses four $2\times 2$ depth-to-space layers, each bit stack
corresponds to a $16\times 16$ image tile. The model uses up to 16
iterations, each of which adds 32 bits to each stack, so the allocation
algorithm makes assignments in steps of 32. % up to a maximum of 512 (pre-entropy
%coding) bits.
We calculate the per-tile quality as the maximum over the mean
$L_1$ error for the four $8 \times 8$ sub-tiles because % we empirically found
%that
averaging over the full $16\times 16$ tile led to visible artifacts for
tiles that span both simple and visually complex image patches.
%Further
%subdivision, however, allocated too many bits to tiles containing small,
%noise-like patches, such as specular highlights.
Finally, we enforce a
heuristic that every tile must use between 50\% and 120\% of the target bit
rate (rounded up to the nearest iteration), to avoid visual artifacts. We
expect that the use of a more accurate perceptual metric would make this
heuristic unnecessary.

Our decoder architecture requires that all bits be present, so we fill missing
bits with a fixed value. Although the network was trained by mapping binary
values to $\pm 1$, we found that using a value of zero led to the best
reconstruction quality. We believe zero works well primarily because the
convolutional layers use zero-padding, which pushes the network to learn that
zero bits are uninformative. Zero is also halfway between the standard bit
values, which can be interpreted as the least biased value, and it is the
absorbing element for multiplication, which will minimize the % absolute
induced response in convolution. % of convolutional kernels outside of the bias term.

SABR requires a small addition to the bitstream generated by our model so that
the decoder knows how many bits are used at each location. This height map is
losslessly compressed using gzip and added to the bitstream. To ensure a fair
comparison, the total size of this metadata is included in all of the bit rate
calculations in \autoref{section: results}.

\subsection{SSIM Weighted Loss}
\label{subsection: ssim}

Training a lossy image compression network introduces a dilemma:
ideally, we would like to train the network using a perceptual metric
as the underlying loss but these metrics are either non-differentiable
or have poorly conditioned gradients. The other option is to use the
traditional $L_1$ or $L_2$ loss; however, these two metrics are only
loosely related to perception. To keep the best of both worlds, we
propose a weighted $L_1$ loss between image $y$ and a reference image $x$
\vspace{-0.5em}
$$L(x, y) = w(x, y) ||y - x||_1, \quad w(x, y) = \frac{S(x,y)}{\bar{S}}$$
\noindent where $S(x,y)$ is a perceptual measure of dissimilarity between
images $x$ and $y$ and where $\bar{S}$ is a dissimilarity baseline.
When doing compression, $y$ is the decompressed version of $x$: $y = f_\theta(x)$
where $\theta$ are the compression model parameters. During training,
the baseline $\bar{S}$ is set to the moving average of $S(y,x)$.
It is not constant but can be considered as almost constant over a short
training window. In our experiments, the moving average decay was $\alpha=0.99$.
To actually perform the gradient update, the trick is to consider the
weight $w(x, y) = \frac{S(x, f_\theta(x))}{\bar{S}}$ as fixed. This
leads to updating using
$\theta' = \theta - \eta w(x, f_\theta(x)) \nabla_\theta ||f_\theta(x) - x||_1$.

Intuitively, this weighted $L_1$ loss is performing dynamic
importance sampling: it compares the perceptual distortion of an image
against the average perceptual distortion and weighs more heavily the
images with high perceptual distortion
%compared to the baseline
and
less heavily the images for which the compression network already
performs well.

In practice, we use a local perceptual measure of dissimilarity. The
image is first split into $8 \times 8$ blocks. Over each of these blocks, a
local weight is computed using $D(x,y) = \frac{1}{2} (1 - SSIM(x,y))$
as the dissimilarity measure (DSSIM), where SSIM refers to the well
known structural similarity index. The loss over the whole image is
then the sum of all these locally weighted losses.
The weighting process can then be thought as a variance minimization
of the perceptual distortion across the image, trying to ensure the
quality of the image is roughly uniform: any $8 \times 8$ block whose
perceptual distortion is higher than the average will be over-weighted
in the loss.

%% file: training.tex
\begin{table}
%  \begin{minipage}{\columnwidth}
  \centering \setstretch{0.95}
  \setlength\tabcolsep{3 pt}
\begin{small}
  \begin{tabular}{|c|ccc|ccc|}
    \hline
    & \multicolumn{3}{c|}{Kodak AUC (dB)} &  \multicolumn{3}{c|}{Tecnick AUC (dB)} \\
    Method & MS-SSIM & SSIM & PSNR & MS-SSIM & SSIM & PSNR \\
    \hline
    Baseline & 32.96 & 19.06 & 59.42 &                           35.49 & 22.35 & 64.16 \\
    DSSIM & 33.43 & 20.17 & 60.46 &                              36.02 & 23.03 & 64.82 \\
    Prime & 33.84 & 20.56 & 60.94 &                              36.34 & 23.29 & 65.19 \\
    Best & \textbf{34.20} & \textbf{21.02} & \textbf{61.40} &    \textbf{36.85} & \textbf{23.67} & \textbf{65.66} \\
    \hline
  \end{tabular}
\caption{\small AUC for MS-SSIM (dB), SSIM (dB), and PSNR across Kodak and
  Tecnick.  Baseline uses \autoref{fig: priming arch}-a  and is trained using
  $L_1$ reconstruction loss. DSSIM also uses \autoref{fig: priming
    arch}-a but is trained using DSSIM reconstruction loss.  Prime uses
  3-priming (similar to \autoref{fig: priming arch}-b) and
  DSSIM training loss.  Best is the same as Prime after more training
  steps.$^{\ref{footnote17}}$  3-priming shows the best results, which then continue to improve with additional training (as shown by Best).}
\label{table: priming}
\end{small}
%  \end{minipage}
\end{table}

%\vspace{-0.5em}
\section{Training}
%\vspace{-0.5em}
\label{section: training}
All experiments use a dataset of a random sampling of 6 million
$1280 \times 720$ images on the web. Each minibatch uses $128 \times
128$ patches  randomly sampled from these images. The Adam
optimizer~\cite{kingma2014} is used with an $\epsilon=1.0$,
$\beta_{1}=0.9$  and a $\beta_{2}=0.999$. All experiments were run
with 10 asynchronous workers on NVIDIA Tesla K80 GPUs and clipping all
gradient norms over $0.5$.

To understand the improvement due to perceptual training metric,
separate from those due to hidden-state refinements, we
%start by training
trained two baseline models (\autoref{fig: priming arch}-a):
one using $L_1$ error for our training loss and the second using our
DSSIM loss.  Both of these models were trained with a learning rate of
0.5 and a batch size of 8, for a total of 3.8M steps.

We then built on the improvements seen with DSSIM training to
investigate the improvements from hidden-state priming (\autoref{fig: priming arch}-b) for 3-priming.
This 3-Prime model as trained in the same way as our two baseline models:
a learning rate of 0.5, a batch size of 8, and a total of 3.8M
training steps.

Finally, we trained additional models (all using DSSIM training)
to investigate $k$-diffusion for $k = 0$ (which is the same as
the DSSIM-trained baseline model), 1, 2, and 3.  For $k=1$ to 3, we
repeat the ``Encoder Diffusion'' and ``Decoder Diffusion'' steps
(\autoref{fig: priming arch}-c) $k$
times before taking the next step's outputs (bits, for the encoder, or
reconstructions, for the decoder) and we do that before every
iteration (not just the first, as in priming).  For a fair comparison
between these models and the DSSIM-trained baseline, we used a
learning rate of 0.2, a batch size of 4, and a total of 2.2M
steps.\footnote{The smaller batch size was needed due to memory
constraints, which forced our learning rate to be lower.}

%% file: results.tex
%\vspace{-0.5em}
\section{Results}
%\vspace{-0.5em}
\label{section: results}

In this section, we first evaluate the performance improvements provided to
our compression architecture, using our proposed techniques: that is,
DSSIM training; priming; and diffusion. Due to the fact that our methods
are intended to preserve color information, the computation of all the metrics
we report is performed in the RGB domain, following~\cite{TwitterComp, Toderici2017cvpr}.

\begin{table}
%  \begin{minipage}{\columnwidth}
  \centering \setstretch{0.95}
  \setlength\tabcolsep{3 pt}
\begin{small}
  \begin{tabular}{|c|ccc|ccc|}
    \hline
    k steps of & \multicolumn{3}{c|}{Kodak AUC (dB)} &  \multicolumn{3}{c|}{Tecnick AUC (dB)} \\
    Diffusion & MS-SSIM & SSIM & PSNR & MS-SSIM & SSIM & PSNR \\
    \hline
    0 & 31.89 & 18.75 & 58.73                             & 34.34 & 21.78 & 63.18 \\
    1 & 33.05 & 19.62 & 59.91                             & 35.41 & 22.52 & 64.23 \\
    2 & 32.85 & 19.38 & 59.81                             & 35.28 & 22.12 & 64.13 \\
    3 & \textbf{33.40} & \textbf{19.87} & \textbf{60.35}  & \textbf{35.68} & \textbf{22.70} & \textbf{64.70} \\
    \hline
  \end{tabular}
\caption{\small AUC for MS-SSIM (dB), SSIM (dB), and PSNR across Kodak and
  Tecnick.  All methods in this table used DSSIM for training loss and
  used diffusion (similar to \autoref{fig: priming arch}-c) with
  different numbers of steps between iterations.$^{\ref{footnote17}}$
  3-diffusion provides the best performance in this test (but at a high computational cost).}
\label{table: diffusion}
\end{small}
%  \end{minipage}
\end{table}

Next, we show the results for the best model that we have trained to date,
which uses 3-priming, trained with DSSIM (but has trained for more steps
than the models used in \autoref{subsection: eval}).  We compare this model
against contemporary image compression codecs (BPG 420; JPEG2000;
WebP; and JPEG) as well as the best recently published
neural-network-based approach~\cite{TwitterComp} and~\cite{Toderici2017cvpr}.

We present results on both Kodak~\cite{kodak} and Tecnick~\cite{tecnick}
datasets.
The Kodak dataset is a set of 24 $768 \times 512$ images (both
landscape and portrait) commonly used as a benchmark for
compression. We also compare using the Tecnick SAMPLING dataset
(100 $1200 \times 1200$ images). We feel
the Tecnick images
are more representative of contemporary, higher resolution content.

\subsection{Comparative Algorithm Evaluation}
\label{subsection: eval}

In this subsection, all of our experiments use nominal bit rates:
neither entropy coding nor SABR were applied to
the RD curves before computing the area under the curve (AUC)
values listed in
Tables~\ref{table: priming} and~\ref{table: diffusion}.

\stepcounter{footnote}
\footnotetext{For Tables~\ref{table: priming} and~\ref{table:
    diffusion}, no entropy compression or SABR was used: these AUC numbers
  can not be compared to those in \autoref{subsection: best eval}. \label{footnote17}}

\begin{figure*}
  \begin{minipage}{\columnwidth}
  \centering \setstretch{0.95}
  \begin{overpic}[width=\linewidth]{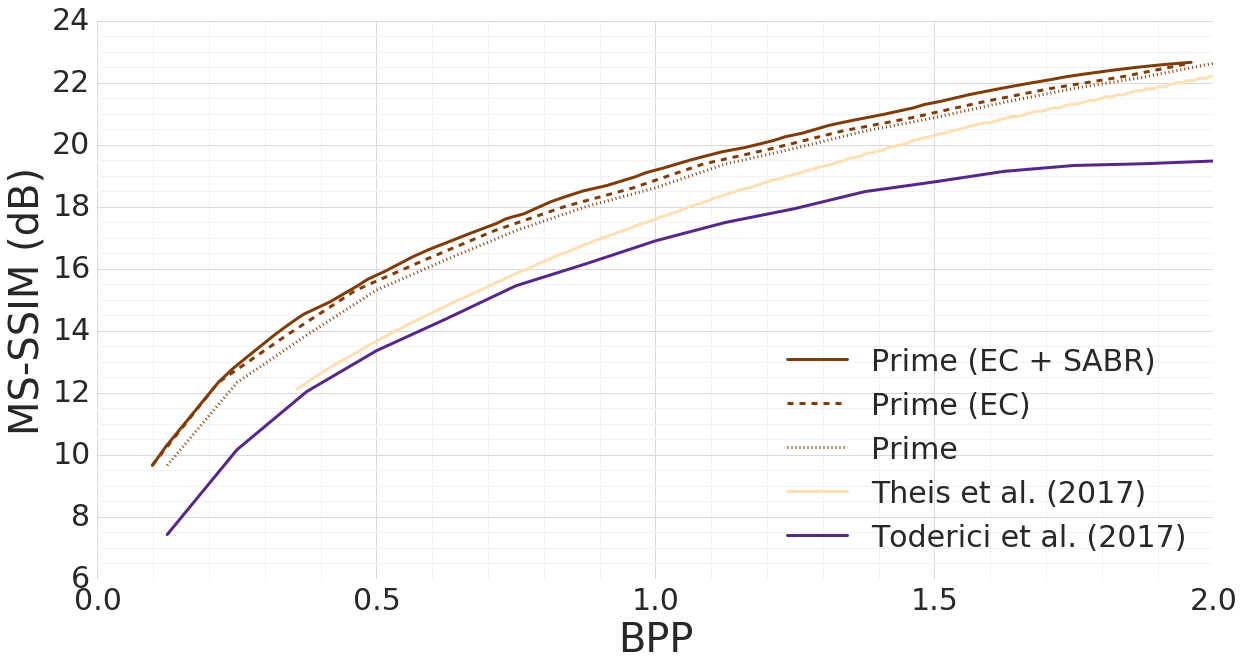}
    \put (10,48) {a)}
  \end{overpic}
    \begin{overpic}[width=\linewidth]{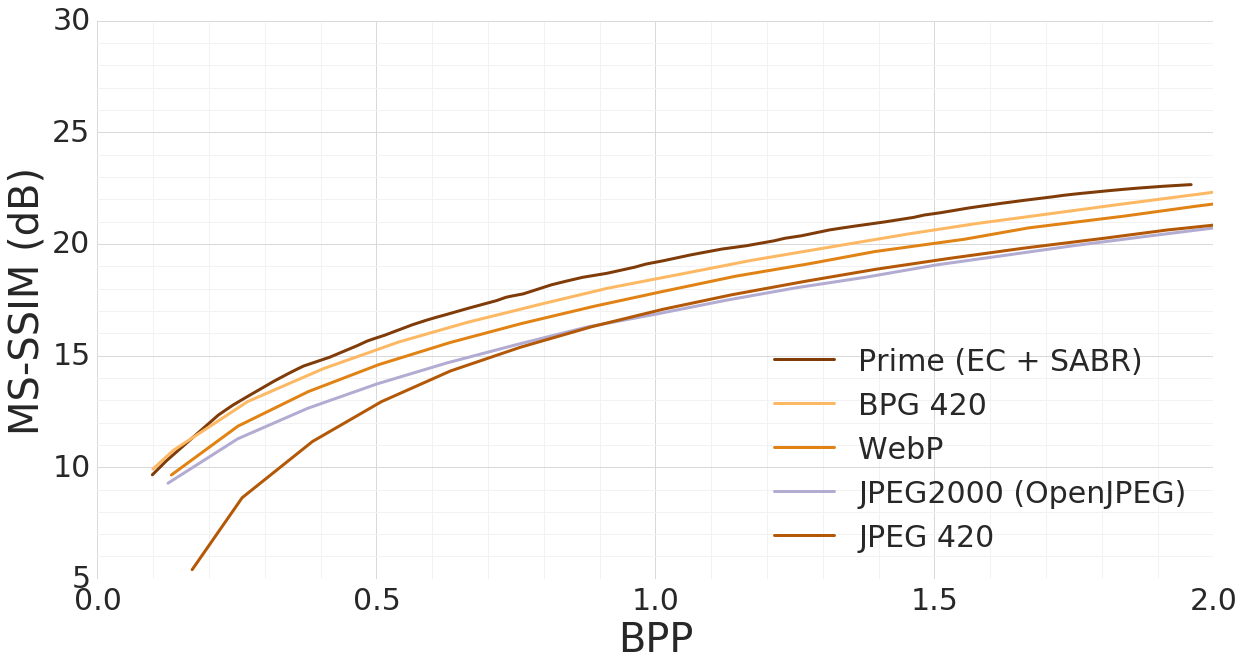}
    \put (10,48) {b)}
  \end{overpic}
    \caption{\small Our full method outperforms existing codecs at all but the
      lowest bit rate where only BPG 420 matches our performance. This figure
      shows MS-SSIM (dB) on Kodak: (a) our method compared
      to~\cite{TwitterComp} and~\cite{Toderici2017cvpr} (without entropy
      coding), and (b) compared to standard image compression codecs.}

  %% \caption{\small MS-SSIM (dB) averaged over Kodak, comparing our
  %% methods (Prime, Prime with entropy coding and Prime with entropy
  %% coding and SABR) with~\cite{TwitterComp} and~\cite{Toderici2017cvpr}
  %% (pre-entropy coded results) (a) and comparing our Prime with entropy
  %% coding and SABR with popular image compression codecs. (b)  Prime
  %% with entropy coding and SABR outperforms all other methods at all
  %% but the lowest bit rate (where BPG 420 matches its performance).}

  \label{fig: msssim_overview}
  \end{minipage}
  \hfill
  \begin{minipage}{\columnwidth}
  \centering \setstretch{0.95}
  \begin{overpic}[width=\linewidth]{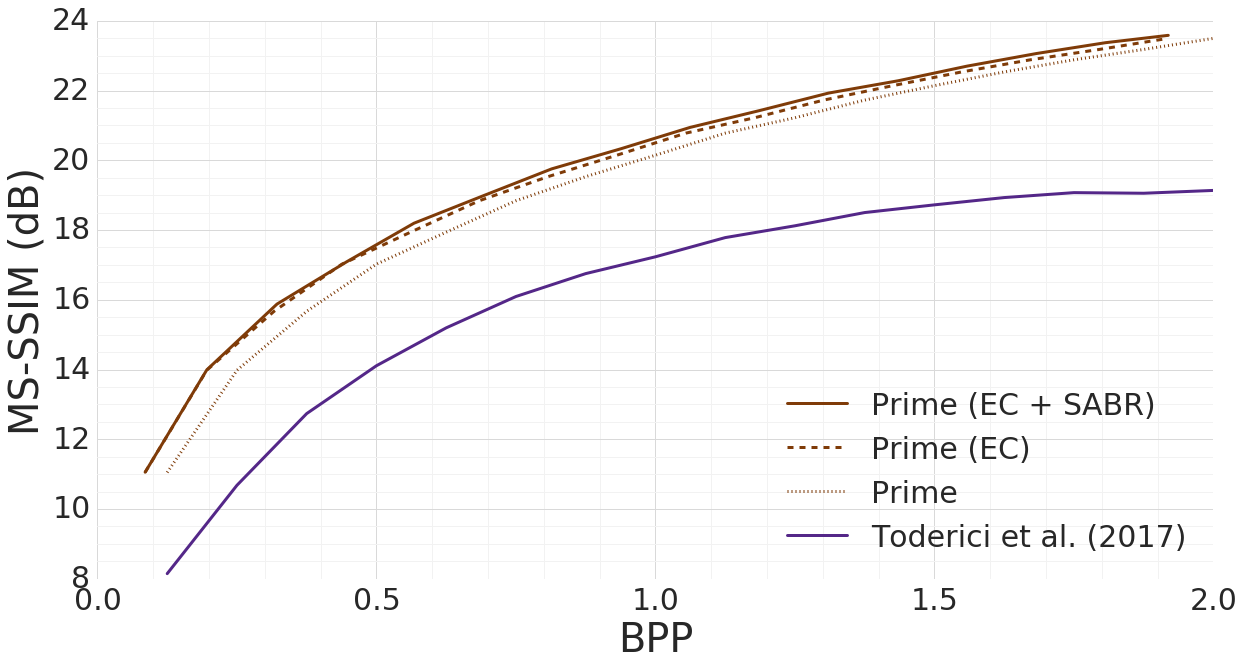}
    \put (10,48) {a)}
  \end{overpic}
  \begin{overpic}[width=\linewidth]{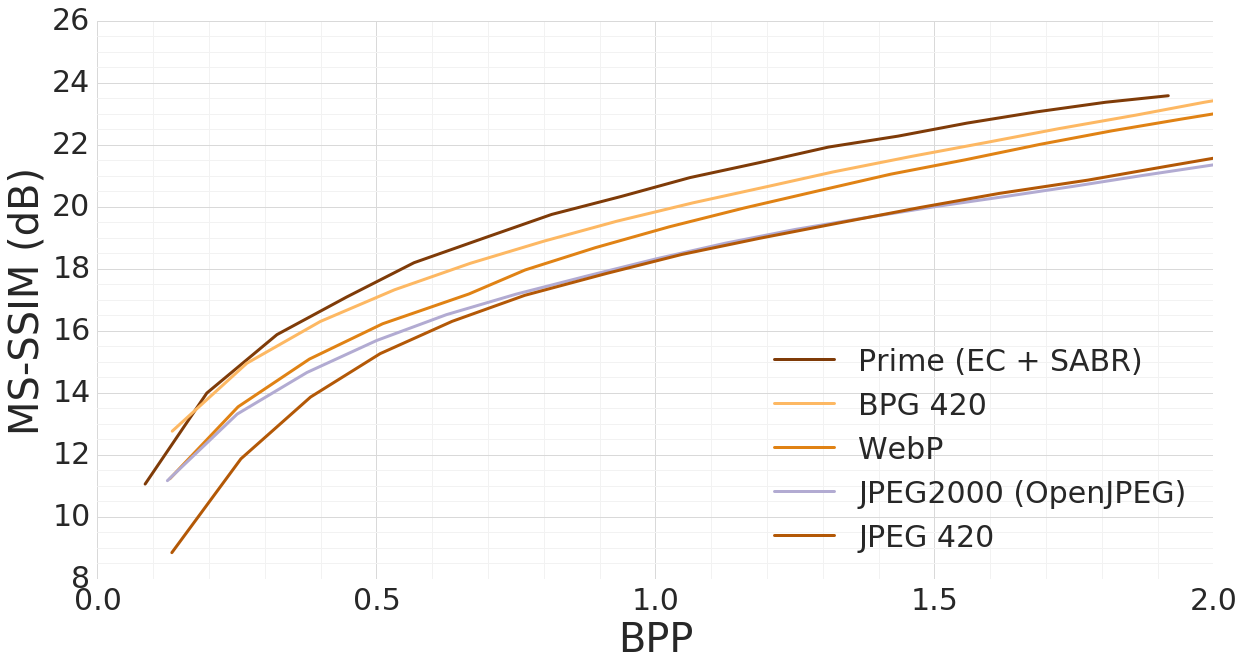}
    \put (10,48) {b)}
  \end{overpic}
  \caption{\small On the larger Tecnick dataset, our full method outperforms
    existing codecs at all but the lowest bit rate where BPG 420 has a small
    advantage. This figure shows MS-SSIM (dB) on Tecnick: (a) our method
    compared to~\cite{Toderici2017cvpr} (results on Tecnick were not available
    for~\cite{TwitterComp}), and (b) compared to standard image codecs.}

  %% \caption{\small MS-SSIM (dB) averaged over Tecnick, comparing our
  %% methods Prime, Prime with entropy coding and Prime with entropy
  %% coding and SABR (a) and comparing our Prime with entropy coding and
  %% SABR with popular image compression codecs. (b) Same as when tested
  %% on lower-resolution images (see Figure~6), Prime
  %% with entropy coding and SABR outperforms all other methods at all
  %% but the lowest bit rate (where BPG 420 matches its performance).}

  \label{fig: msssim_overview_tecnick}
  \end{minipage}
\end{figure*}

We evaluate our results using AUC for
peak signal-to-noise ratio (PSNR), SSIM (dB) and MS-SSIM (dB).
SSIM (dB) and MS-SSIM (dB) are $-10 \log_{10}(1 - Q)$ where $Q$ is
either SSIM~\cite{ssim} or MS-SSIM~\cite{wang2003multiscale}.
Both of these metrics tend to have significant quality
differences in the range between 0.98 and 1.00, making them difficult
to see on linear graphs and difficult to measure with AUC.  This dB transform
is also supported by the original MS-SSIM~\cite{wang2003multiscale} publication,
which showed the mean opinion score is linearly correlated with the MS-SSIM score
after transforming that score to the log domain. Subsequent compression
studies have also adopted this convention, if the methods were able
to achieve high-quality--compression results~\cite{WebP}.

The Baseline and DSSIM models differ only in the training loss that
was used ($L_1$ or DSSIM).  As shown in \autoref{table:
priming}, the DSSIM model does better
(in terms of AUC) for all of the metrics on both image test sets.
Surprisingly, this
is true even of PSNR, even though the $L_1$ loss function should be
closer to PSNR than the DSSIM-weighted $L_1$ loss.
The Prime model (trained with DSSIM loss) does better than the
non-priming model, even when both are evaluated at the same number of
training steps (``Prime'' versus ``DSSIM'' in \autoref{table:
  priming}).  The Prime model continues to improve with additional
training, as shown by the results labeled ``Best'' in \autoref{table:
  priming}.  While the runtime computation is
increased by the use of priming, the percent increase is limited since
these extra steps only happen before the first iteration (instead of
between all iterations, as with diffusion).

\autoref{table: diffusion} reports our AUC results on a second set of models,
comparing different numbers of diffusion steps (extensions of
\autoref{fig: priming arch}-c). The results from
this experiment show that more diffusion (up to the 3 we tested)
increases reconstruction quality.  However, as the number of diffusion steps
goes up, the resources used also increases: for a $k$-diffusion
network, compression/decompression computation goes up nearly linearly
with $k+1$.  Training time is also substantially longer for increasing
$k$.  In light of these practical trade offs, we have focused on
the Prime model for our comparisions in \autoref{subsection: best eval}

\begin{figure*}
  \centering
  \begin{minipage}{.985\textwidth}
  \centering \setstretch{0.8}
  \begin{tabular}{cccc}
    \small{JPEG2000} & \small{WebP} & \small{BPG 420} & \small{Our Method} \\
    \includegraphics[width=0.22\textwidth]{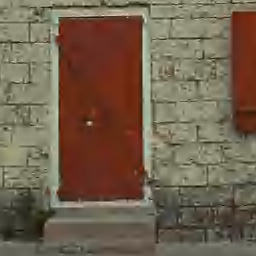} &
    \includegraphics[width=0.22\textwidth]{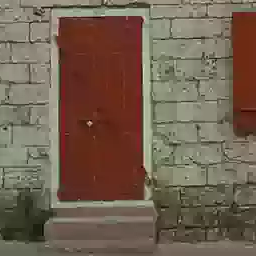} &
    \includegraphics[width=0.22\textwidth]{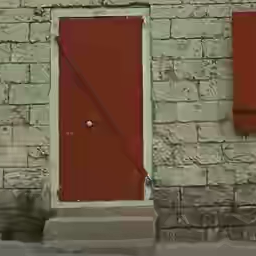} &
    \includegraphics[width=0.22\textwidth]{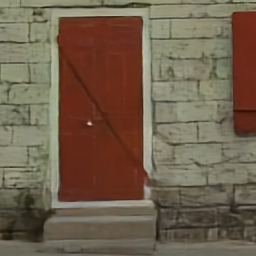} \\
    \small{0.250 bpp} & \small{0.252 bpp} & \small{0.293 bpp} & \small{0.234 bpp} \\
    
    \includegraphics[width=0.22\textwidth]{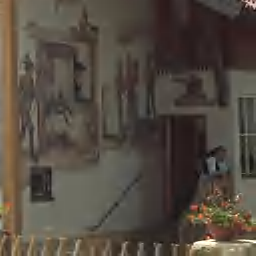} &
    \includegraphics[width=0.22\textwidth]{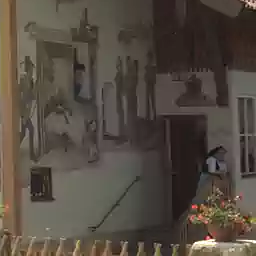} &
    \includegraphics[width=0.22\textwidth]{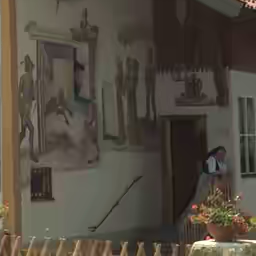} &
    \includegraphics[width=0.22\textwidth]{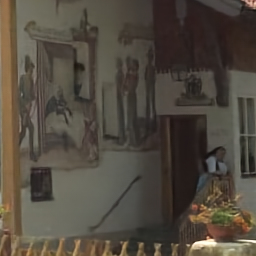} \\   
    \small{0.502 bpp} & \small{0.504 bpp} & \small{0.504 bpp} & \small{0.485 bpp} \\
    
    \includegraphics[width=0.22\textwidth]{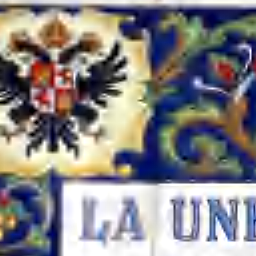} &
    \includegraphics[width=0.22\textwidth]{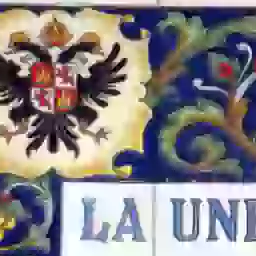} &
    \includegraphics[width=0.22\textwidth]{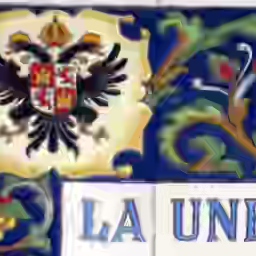} &
    \includegraphics[width=0.22\textwidth]{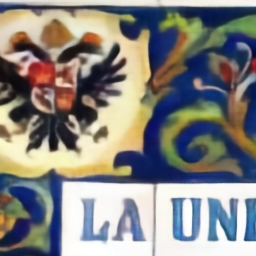} \\
     \small{0.125 bpp} & \small{0.174 bpp} & \small{0.131 bpp} & \small{0.122 bpp} \\
    
    \includegraphics[width=0.22\textwidth]{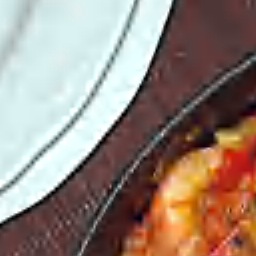} &
    \includegraphics[width=0.22\textwidth]{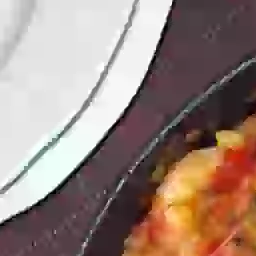} &
    \includegraphics[width=0.22\textwidth]{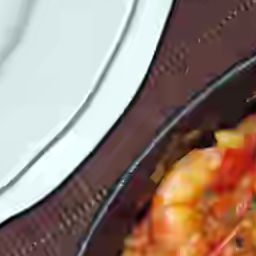} &
    \includegraphics[width=0.22\textwidth]{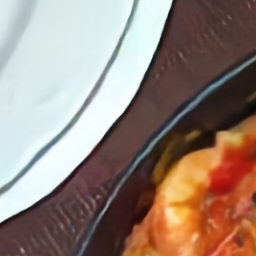} \\
     \small{0.125 bpp} & \small{0.131 bpp} & \small{0.125 bpp} & \small{0.110 bpp} \\
    
    \includegraphics[width=0.22\textwidth]{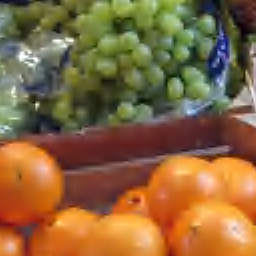} &
    \includegraphics[width=0.22\textwidth]{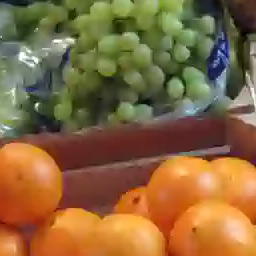} &
    \includegraphics[width=0.22\textwidth]{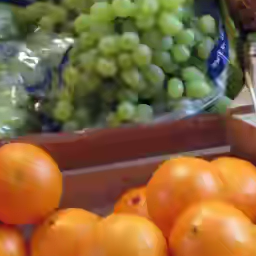} &
    \includegraphics[width=0.22\textwidth]{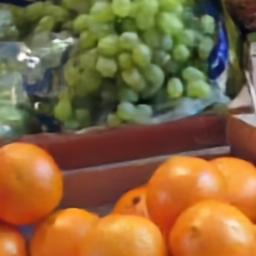} \\
    \small{0.250 bpp} & \small{0.251 bpp} & \small{0.251 bpp} & \small{0.233 bpp} \\
  \end{tabular}
  \end{minipage}
\vspace{0.3em}

  \caption{\small Example patches comparing our Best-model
    results with BPG 420, JPEG2000 (OpenJPEG) and WebP.
    % at different bit rates
    For the most visible differences, consider: (first row) the cross bar on door;
    (second row) the handrail and the hanging light in front of the dark wood;
    (third row) the text; (fourth
    row) the pan edge and the plate rim; (fifth row) the outlines of the oranges
    and the crate edge.}
  \label{fig: example images}
\end{figure*}

\subsection{Best Results Evaluation}
\label{subsection: best eval}

The Prime model trained for 4.1 million steps is our best model to date
(called ``Best'' in the tables above). This section evaluates the results when
adding entropy coding and SABR to this model.

In \autoref{fig: msssim_overview}-a, we compare our best model, with
and without entropy coding, to the work reported by Theis et
al.~\cite{TwitterComp}. For our entropy coding we train the
probability modeler, described in~\cite{Toderici2017cvpr}, using
the codes generated by our model operating on the set of 6
million web images, mentioned in \autoref{section: training}.

Figures~\ref{fig: msssim_overview}-a and~\ref{fig: msssim_overview_tecnick}-a
also show our results using SABR (in conjunction with entropy coding) to
obtain even higher compression rates.  It should be noted that we do
not retrain the compression model (or the entropy-coding model) to handle SABR: we
use the previously trained models unchanged.  This is an area in which
we could expect even better performance from our model, if we did some
amount of retraining for SABR.

\begin{table}
\begin{small}
  \centering \setstretch{0.95}
  \setlength\tabcolsep{3 pt}
  \resizebox{\linewidth}{!}{
  \begin{tabular}{|c|ccc|ccc|}
    \hline
    & \multicolumn{3}{c|}{Kodak Rate Difference \%} &  \multicolumn{3}{c|}{Tecnick Rate Difference \%} \\
    Method & MS-SSIM & SSIM & PSNR & MS-SSIM & SSIM & PSNR \\
    \hline
    Prime (EC + SABR) & \textbf{43.17} & 39.97 & 27.14 & \textbf{45.65} & 40.08 & 17.36 \\
    Prime (EC) & 41.70 & 36.51 & 19.29 & 44.57 & 36.82 & 9.73 \\
    BPG 444 & 40.04 & 44.86 & \textbf{56.30} & 44.10 & \textbf{44.25} & \textbf{55.54} \\
    BPG 420 & 37.04 & \textbf{46.94} & 54.85 & 36.27 & 43.02 & 48.68 \\
    Prime & 36.32 & 30.89 & 12.20 & 35.05 & 26.86 & -6.09 \\
    JPEG2000 (Kakadu) & 31.75 & 22.23 & 28.29 & 35.18 & 27.44 & 27.08 \\
    WebP & 26.85 & 29.85 & 36.33 & 24.28 &  23.35 & 23.14 \\
    JPEG2000 (OpenJPEG) & 15.99 & 24.80 & 38.28 & 14.34 & 20.70 & 26.08 \\
    Theis et al.\cite{TwitterComp} & 15.10 & 28.69 & 29.04 & -- & -- & -- \\
    Toderici et al.\cite{Toderici2017cvpr} & 12.93 & -1.86 & -13.34 & -25.19 & -44.98 & -67.52 \\
    \hline
  \end{tabular}
	}
\end{small}
  \hfill
  \centering \setstretch{0.95}
\caption{\small Bjøntegaard rate-difference on MS-SSIM, SSIM and PSNR for Kodak and Tecnick
  datasets. This shows the
  bit rate difference across each metric (larger numbers are better). Codecs
  are sorted in order of MS-SSIM bit-rate difference, while the best result in
  each metric is bolded.}
\label{table: bjontegaard_kodak}
\end{table}

\begin{figure}
  \centering \setstretch{0.95}
  \begin{overpic}[width=\linewidth]{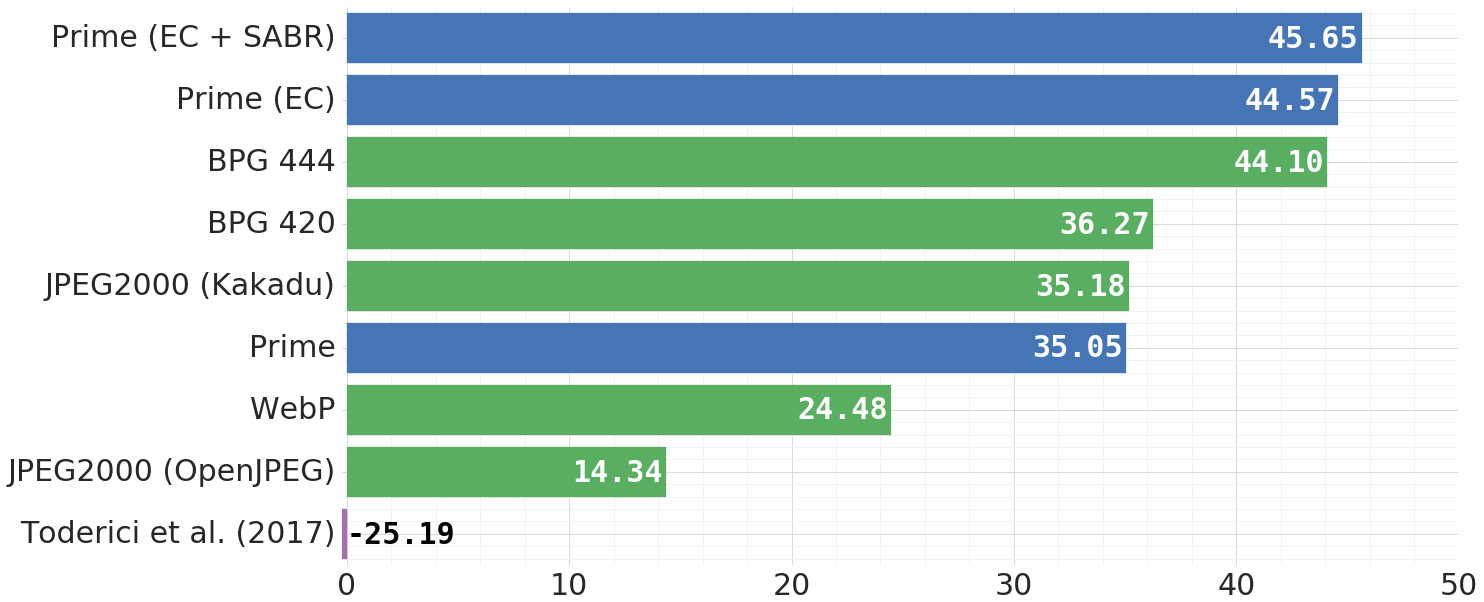}
    \put (84,5) {Tecnick}
  \end{overpic}
  \caption{\small Rate savings (Bj{\o}ntegaard Delta) relative to JPEG
    under MS-SSIM for the Tecnick dataset. By this measure, two of our
      approaches outperform current standard codecs and all of our
  approaches outperform  previous research in network-based codecs.}
  \label{fig: bjontegaard_tecnick_rate}
\end{figure}

Compared to neural network-based methods, our best model has a better MS-SSIM
RD curve than~\cite{TwitterComp, Toderici2017cvpr}.  Our model's curve improves
with entropy coding and improves further with SABR.

In Figures~\ref{fig: msssim_overview}-b and~\ref{fig: msssim_overview_tecnick}-b,
we compare our best model against
many popular image compression codecs. We provide examples of our
compression results, and those of other popular codecs, in
\autoref{fig: example images}.\footnote{Full-image examples are
available in Supplementary Materials.}
For
these image examples, since each of the codecs allows only coarse-level
control of the output bit rate, we bias our comparisons against our own
model.  Specifically, when doing our comparisons, we always choose an
average bit rate that is the same or larger than the bit rate produced
by our method (giving an advantage to the other methods in the comparison).
%The differences between these compression approaches is often most visible
%in areas on or near diagonal edges (e.g., the cross bar on the first row
%and the handrail on the second) or sharp boundaries (e.g., the text in the third
%row and the outline of the individual oranges in the fifth).

Finally, Table~\ref{table: bjontegaard_kodak} and Figures~\ref{fig: bjontegaard_kodak_rate} and~\ref{fig:
bjontegaard_tecnick_rate} summarize the rate
savings of the codecs we have introduced (or compared ourselves to),
using Bj{\o}ntegaard Delta (BD) rate differences~\cite{bjontegaard2001calcuation}.
BD rate differences are
the percent difference in area between two
RD curves, after a logarithmic transform of the bit rate. When
computing BD rate savings on methods that fail to deliver 
the full quality range, 
the difference in area is only computed across quality levels provided
  by both curves. BD rate differences use the log bit rate since
the human visual system is more sensitive
to low-bit-rate areas than to the high-bit-rate areas.\footnote{The Supplementary Materials
  provide additional details about computing Bj{\o}ntegaard measures,
  as well as the quality-improvement BD results.}
%Moreover, the RD
%curves are typically not sampled at the exact same bit rates for all
%methods, so before computing the difference in area, Bj{\o}ntegaard
%also proposed using a polynomial fit of the curves, then sampling the
%fitted curve over 100 points (common for both curves), and using the
%trapezoidal integration method to compute the areas.
The BD difference was
originally defined for PSNR, but since its publication, better
measures of quality have been proposed~\cite{wang2003multiscale}. As a
result, we are reporting the BD rate computed on the
logarithmic transform of MS-SSIM.

% removed RD curves for DSSIM and Prime wo/ entropy coding
%\begin{figure}
%  \centering
%  \includegraphics[width=\linewidth]{images/msssim_kodak_exp.png}
%  \caption{\small Plot showing the reconstruction quality on the DSSIM
%    and Prime experiments in MS-SSIM and averaged over the Kodak
%    dataset.  No entropy coding or SABR were applied to these bit rates.}
%  \label{fig: kodak_exp}
%\end{figure}
%
%\begin{figure}
%  \centering
%  \includegraphics[width=\linewidth]{images/msssim_tecnick_exp.png}
%  \caption{\small Plot showing the reconstruction quality on the DSSIM
%    and Prime experiments in MS-SSIM and averaged over the Tecnick
%    dataset.  No entropy coding or SABR were applied to these bit
%    rates.}
%  \label{fig: tecnick_exp}
%\end{figure}

% removed RD curves for Diffusion (with DSSIM) wo/ entropy coding
%\begin{figure}
%  \centering
%  \includegraphics[width=\linewidth]{images/msssim_kodak_abp.png}
%  \caption{\small Plot showing the reconstruction quality on the
%    diffusion experiments in MS-SSIM and averaged over the Kodak
%    dataset. No entropy coding or SABR were applied to these bit
%    rates.}
%  \label{fig: kodak_abp}
%\end{figure}
%
%\begin{figure}
%  \centering
%  \includegraphics[width=\linewidth]{images/msssim_tecnick_abp.png}
%  \caption{\small Plot showing the reconstruction quality on the
%    diffusion experiments in MS-SSIM and averaged over the Tecnick
%    dataset.  No entropy coding or SABR were applied to these bit
%    rates.}
%  \label{fig: tecnick_abp}
%\end{figure}

%% file: conclusion.tex
%\vspace{-0.5em}
\section{Conclusion}
%\vspace{-0.5em}
\label{section: conclusion}
We introduced three techniques --- perceptually-weighted training loss,
hidden-state priming, and spatially adaptive bit rates --- and showed that
they boost the performance of our baseline recurrent image compression
architecture. Training with a perceptually-weighted $L_1$ loss improved our model's
performance on MS-SSIM, SSIM, and PSNR. Hidden-state priming provides further
improvements to reconstruction quality, similar to that of diffusion but with
lower computational requirements during inference and training. The quality
improvements seen with priming are probably related to initializing the hidden
states within the network with content-dependent values.

The additional quality improvements seen with diffusion are probably related
to the increased spatial context available in the decoder: with 3-diffusion,
the decoder's context nearly doubles, increasing from about $6 (t+1)$ to $10
(t + 1)$ ``bit stacks'' (where a bit stack occurs each $16 \times 16$ pixels). Finally, SABR can reduce the bit rate in the areas of the image that are
easier to compress. In our models, this adaptivity more than makes up for the
additional overhead needed to send the SABR height map.

 %% The resulting model performs better than existing image codecs (BPG 420,
 %% WebP, JPEG2000, and JPEG) as well as recent neural-network-based compression
 %% models (\cite{TwitterComp} and~\cite{Toderici2017cvpr}) on MS-SSIM.

Combining these three techniques improves the MS-SSIM rate-distortion curve
for our GRU-based architecture, surpassing that of recent neural-network-based
methods (\cite{TwitterComp} and~\cite{Toderici2017cvpr}) and many standard
image codecs (BPG 420, WebP, JPEG2000, and JPEG). Our approach is the first
neural-network-based codec shown to outperform WebP and provide competitive
coding efficiency with some variants of BPG.

%% This combination also leaves open many areas of research,
%% including better SABR strategies and  more flexible Priming (or
%% Diffusion methods).